\documentclass[10pt,twocolumn,letterpaper]{article}

\usepackage{iccv}              

\usepackage{listings}
\usepackage{algorithm}
\usepackage{algpseudocode}

\usepackage{multirow}
\usepackage{rotating}
\usepackage{graphicx}
\usepackage{siunitx}
\usepackage{makecell}
\usepackage{bm}
\usepackage{amsmath}
\usepackage{gensymb}
\usepackage{siunitx}
\usepackage{enumitem}
\usepackage{colortbl}%

\makeatletter
\g@addto@macro\normalsize{%
  \setlength\abovedisplayskip{3pt}
  \setlength\belowdisplayskip{3pt}
  \setlength\abovedisplayshortskip{0pt}
  \setlength\belowdisplayshortskip{0pt}
}
\makeatother

\definecolor{lightbrown}{rgb}{0.996,0.910,0.784}
\definecolor{midbrown}{rgb}{0.996,0.816,0.647}
\definecolor{darkbrown}{rgb}{0.992,0.733,0.518}
\definecolor{iccvblue}{rgb}{0.21,0.49,0.74}
\usepackage[pagebackref,breaklinks,colorlinks,allcolors=iccvblue]{hyperref}

\def\paperID{1490} 
\def\confName{ICCV}
\def\confYear{2025}

\title{7DGS: Unified Spatial-Temporal-Angular Gaussian Splatting}

\author{Zhongpai Gao\thanks{Corresponding author}, Benjamin Planche, Meng Zheng, Anwesa Choudhuri, Terrence Chen, Ziyan Wu \\
United Imaging Intelligence, Boston MA, USA \\
\texttt{\{first.last\}@uii-ai.com}\\
}

\begin{document}
\twocolumn[{%
\renewcommand\twocolumn[1][]{#1}%
\maketitle
\begin{center}
\vspace{-2.5em}
  \centering
    \captionsetup{type=figure}
    \includegraphics[width=0.99\textwidth]{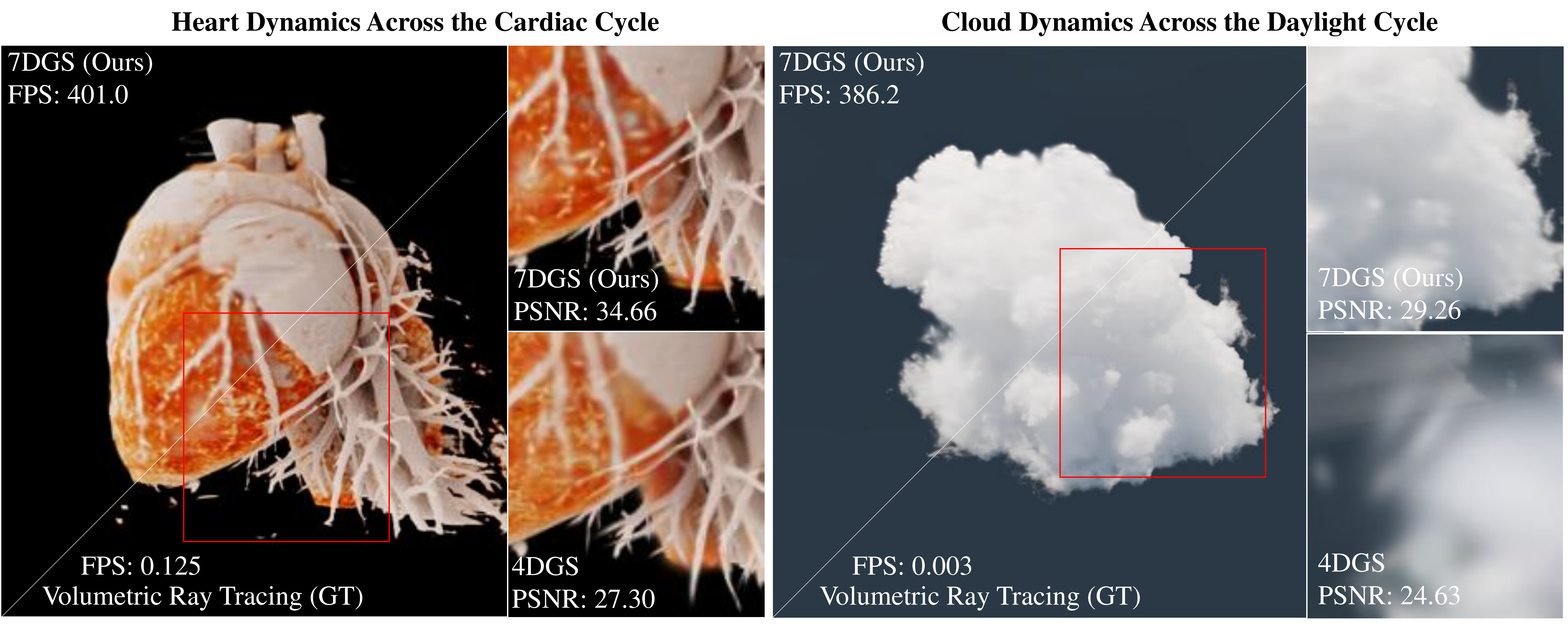}
    \vspace{-0.5em}
    \captionof{figure}{Visualization of volumetric rendering for dynamic scenes. \textbf{Top-left:} Our 7DGS rendering. \textbf{Bottom-left:} Physically-based rendering via ray/path tracing (note: floating artifacts in the \texttt{heart} scene are caused by incomplete segmentation in CT scans and are not rendering artifacts). \textbf{Right:} Comparison between our method and 4DGS in highlighted red regions.}
   \label{fig:teaser}
\end{center}
}]

\begin{abstract}
Real-time rendering of dynamic scenes with view-dependent effects remains a fundamental challenge in computer graphics. While recent advances in Gaussian Splatting have shown promising results separately handling dynamic scenes (4DGS) and view-dependent effects (6DGS), no existing method unifies these capabilities while maintaining real-time performance. We present 7D Gaussian Splatting (7DGS), a unified framework representing scene elements as seven-dimensional Gaussians spanning position (3D), time (1D), and viewing direction (3D). Our key contribution is an efficient conditional slicing mechanism that transforms 7D Gaussians into view- and time-conditioned 3D Gaussians, maintaining compatibility with existing 3D Gaussian Splatting pipelines while enabling joint optimization. Experiments demonstrate that 7DGS outperforms prior methods by up to 7.36 dB in PSNR while achieving real-time rendering (401 FPS) on challenging dynamic scenes with complex view-dependent effects. The project page is: \href{https://gaozhongpai.github.io/7dgs/}{gaozhongpai.github.io/7dgs/}.
\end{abstract}

\section{Introduction}

Photorealistic rendering of dynamic scenes with complex view-dependent effects remains challenging in computer vision and graphics. Examples include dynamic heartbeat visualization from real CT scans and clouds transitioning across daylight with absorption and scattering effects (Figure \ref{fig:teaser}). The ability to synthesize novel views of dynamic scenes is crucial for numerous applications, including virtual reality, augmented reality, content creation, and digital twins. While significant progress has been made in static scene reconstruction and rendering through Neural Radiance Fields (NeRF) \cite{mildenhall2020nerf} and more recently through 3D Gaussian Splatting (3DGS) \cite{kerbl20233dgs}, achieving high-quality, real-time rendering of dynamic scenes with view-dependent effects presents substantial computational and representational challenges.

The core difficulty lies in simultaneously modeling three fundamental aspects: 1) spatial geometry, 2) temporal dynamics, and 3) view-dependent appearance. Each of these dimensions introduces unique challenges. Spatial modeling must capture intricate scene geometry at varying scales. Temporal modeling must represent both rigid and non-rigid motions with potentially complex deformations. View-dependent modeling needs to capture sophisticated light transport effects such as scattering, anisotropic reflections, and translucency. When considered together, these challenges become significantly more complex due to their interdependencies—for instance, specular highlights on moving objects change their appearance based on both viewing direction and object position over time.

Recent advances have addressed these challenges in isolation. 3DGS \cite{kerbl20233dgs} introduced a breakthrough in static scene rendering by representing scenes as collections of 3D Gaussian primitives, enabling real-time rendering rates while maintaining high visual fidelity. However, this approach is inherently limited to static scenes. Two recent extensions have independently addressed different limitations of 3DGS: 4D Gaussian Splatting (4DGS) \cite{yang2023real} incorporates temporal dynamics by extending the representation to 4D (space+time), while 6D Gaussian Splatting (6DGS) \cite{gao20246dgs} models view-dependent effects by adding directional dimensions (space+direction). Despite their success in their respective domains, neither approach provides a comprehensive solution for dynamic scenes with view-dependent effects, as they address only subsets of the challenge.

In this paper, we present 7D Gaussian Splatting (7DGS), a unified framework for real-time rendering of dynamic scenes with view-dependent effects. Our key insight is to model scene elements as 7-dimensional Gaussians spanning spatial position (3D), time (1D), and viewing direction (3D). This high-dimensional representation naturally captures the interdependencies between geometry, dynamics, and appearance, enabling more accurate modeling of complex phenomena such as moving specular highlights and time-varying anisotropic reflections.

The primary technical challenge in our approach is efficiently handling 7D Gaussians while maintaining real-time performance. To address this, we introduce a principled conditional slicing mechanism that transforms 7D Gaussians into time- and view-conditioned 3D Gaussians compatible with existing real-time rendering pipelines. This operation preserves the computational efficiency of 3DGS while incorporating the rich representational capacity of our 7D model. Furthermore, we develop an adaptive Gaussian refinement technique that dynamically adjusts Gaussian parameters via neural network-predicted residuals, enabling more accurate modeling of complex non-rigid deformations and time-varying appearance. 

We evaluate 7DGS on two public datasets: D-NeRF \cite{pumarola2021d} (synthetic monocular videos) and Technicolor \cite{sabater2017dataset} (in-the-wild multi-view videos), and a custom dataset 7DGS-PBR with dynamic scenes featuring complex motions and view-dependent effects. Our results demonstrate that 7DGS consistently outperforms existing methods in terms of both rendering quality and computational efficiency. Our contributions can be summarized as follows:
\begin{itemize}
    \item \textbf{Unified High-Dimensional Representation:} We introduce a novel 7D Gaussian model that jointly encodes spatial structure, temporal evolution, and view-dependent appearance. Furthermore, an adaptive Gaussian refinement technique is developed to enable more accurate modeling of complex deformations and time-varying appearance.
    \item \textbf{Efficient Conditional Slicing:} By deriving a principled conditional slicing mechanism, our method projects high-dimensional Gaussians into 3D counterparts that are compatible with existing real-time rendering pipelines, ensuring both efficiency and fidelity.
    \item \textbf{Validation:} Extensive experiments demonstrate that 7DGS outperforms the prior method 4DGS by up to 7.36 dB in PSNR while maintaining real-time rendering speeds (exceeding 401 FPS) on challenging dynamic scenes exhibiting complex view-dependent effects.
\end{itemize}

\section{Related Work}

\noindent\textbf{Dynamic Neural Radiance Fields.} 
NeRF~\cite{mildenhall2020nerf} revolutionized novel view synthesis by representing scenes as continuous volumetric functions parameterized by neural networks. While the original NeRF focused on static scenes, numerous extensions \cite{guo2023forward, liu2023robust, wang2023flow, johnson2023unbiased, fang2022fast, lou2024darenerf, song2022pref, yan2023nerf} have emerged for dynamic scene modeling. D-NeRF~\cite{pumarola2021d}, Nerfies~\cite{park2021nerfies}, and HyperNeRF~\cite{park2021hypernerf} condition on time and learn deformation fields that warp points from canonical space to each time step. DyNeRF~\cite{li2022neural} represents scene dynamics using compact latent codes with a time-conditioned neural radiance field. To improve efficiency, HexPlane~\cite{cao2023hexplane} accelerated dynamic NeRF rendering through hybrid representations. Despite these advances, NeRF-based methods generally struggle to achieve real-time performance when modeling complex dynamics and view-dependent effects.

\vspace{0.5em}
\noindent\textbf{Dynamic 3D Gaussian Splatting.} 3DGS~\cite{kerbl20233dgs} represents scenes as collections of 3D Gaussians with learnable parameters, enabling high-quality rendering at real-time rates through efficient rasterization. Building on this foundation, several works \cite{yang2024deformable, lin2024gaussian, lu20243d, huang2024sc, qian20243dgs} have extended 3DGS for dynamic scenes. 4DGS~\cite{yang2023real} incorporates temporal dynamics by extending Gaussians to a 4D (space+time) representation. Dynamic 3D Gaussians~\cite{luiten2023dynamic} and 4D Gaussians~\cite{wu20244d} jointly optimize Gaussians in canonical space alongside a deformation field to model scene geometry and dynamics. Ex4DGS \cite{lee2024fully} explicitly models the motions of 3D Gaussians using keyframe interpolation. While these approaches successfully address temporal aspects of dynamic scene modeling, they do not fully account for view-dependent effects within a unified framework.

\vspace{0.5em}
\noindent\textbf{View-dependent Rendering.}
For view-dependent effects, various methods have incorporated sophisticated physically-based reflectance models into neural rendering pipelines. NeRV~\cite{srinivasan2021nerv} introduced neural reflectance and visibility fields to capture view-dependent appearance. LFNR~\cite{suhail2022light} proposed light field neural rendering for realistic view synthesis, while PhySG~\cite{zhang2021physg} incorporated physically-based BRDF models. In parallel, 6DGS~\cite{gao20246dgs} extended 3DGS to capture rich angular variations through 6D (space+direction) Gaussians. Recent work \cite{condor2024dontsplatgaussiansvolumetric, blanc2024raygauss, moenne20243d, zhou2024unified, mai2024ever, govindarajan2025radiant} has also focused on integrating Gaussian primitives with ray tracing for more accurate light transport.

Our 7DGS method builds upon these prior works by unifying spatial, temporal, and angular dimensions into a single coherent framework. Unlike previous approaches that address temporal dynamics and view-dependent effects separately, 7DGS jointly models these dimensions through a unified 7D Gaussian representation, capturing their interdependencies while maintaining real-time performance.

\section{Preliminary}

In this section, we review two foundational methods that form the basis of our 7D Gaussian Splatting (7DGS) framework: 3D Gaussian Splatting (3DGS) \cite{kerbl20233dgs} for static scene rendering, and its extension, 6D Gaussian Splatting (6DGS) \cite{gao20246dgs}, which incorporates view-dependent effects.

\vspace{0.5em}
\noindent\textbf{3D Gaussian Splatting.}
3DGS represents a scene as a collection of anisotropic 3D Gaussians. Each Gaussian is defined by a mean vector 
\(
\mu \in \mathbb{R}^3,
\)
which specifies its spatial position, and a covariance matrix 
\(
\Sigma \in \mathbb{R}^{3 \times 3},
\)
which encodes the extent, shape, and orientation of the Gaussian. In practice, the covariance is factorized as
\begin{align}
\Sigma = R\,S\,R^\top,
\end{align}
where \(S = \operatorname{diag}(s_x,s_y,s_z)\) is a diagonal scaling matrix and \(R\) is a rotation matrix that aligns the Gaussian with the global coordinate system. This factorization provides an intuitive and compact way to represent local geometry.

In addition to geometry, each Gaussian carries an opacity \(\alpha\) and view-dependent color information. The color is modeled via spherical harmonics:
\begin{align}
c(d) = \sum_{\ell=0}^{N} \sum_{m=-\ell}^{\ell} \beta_{\ell m} Y_{\ell m}(d),
\end{align}
where $N$ is the harmonics order ($N=3$ typically), \(d\) denotes the viewing direction, \(\beta_{\ell m}\) are learnable coefficients, and \(Y_{\ell m}(d)\) are the spherical harmonic basis functions. This representation enables the model to capture complex appearance variations under different viewing angles while maintaining real-time rendering capabilities through efficient rasterization.

\vspace{0.5em}
\noindent\textbf{6D Gaussian Splatting.}
While 3DGS excels at static scene rendering, it does not account for the appearance changes induced by view-dependent effects. To overcome this limitation, 6D Gaussian Splatting extends the 3D representation by incorporating directional information. In 6DGS, each scene element is modeled as a 6D Gaussian defined over a joint space:
\begin{align}
X = \begin{pmatrix} X_p \\ X_d \end{pmatrix} \sim \mathcal{N}\!\left(
\begin{pmatrix}
\mu_p \\ \mu_d
\end{pmatrix},
\begin{pmatrix}
\Sigma_p & \Sigma_{pd} \\
\Sigma_{pd}^\top & \Sigma_d
\end{pmatrix}
\right).
\end{align}
Here, \(X_p \in \mathbb{R}^3\) represents the spatial coordinates with mean \(\mu_p\) and covariance \(\Sigma_p\), while \(X_d \in \mathbb{R}^3\) encodes the directional component with mean \(\mu_d\) and covariance \(\Sigma_d\). The cross-covariance \(\Sigma_{pd}\) captures correlations between position and direction, allowing the Gaussian to encode view-dependent appearance variations.

For numerical stability and to guarantee positive definiteness, the full 6D covariance is parameterized via a Cholesky decomposition:
\begin{align}
\Sigma = LL^\top,
\end{align}
with \(L\) being a lower-triangular matrix whose diagonal entries are enforced to be positive. To render an image for a given viewing direction \(d\), the 6D Gaussian is conditioned on \(X_d = d\), yielding a conditional 3D Gaussian for the spatial component. Specifically, the conditional distribution is given by:
\begin{align}
p(X_p \mid X_d = d) \sim \mathcal{N}(\mu_{\text{cond}}, \Sigma_{\text{cond}}),
\end{align}
with
\begin{align}
\mu_{\text{cond}} &= \mu_p + \Sigma_{pd}\,\Sigma_d^{-1}(d-\mu_d), \\
\Sigma_{\text{cond}} &= \Sigma_p - \Sigma_{pd}\,\Sigma_d^{-1}\,\Sigma_{pd}^\top.
\end{align}
Moreover, the opacity of each Gaussian is modulated to reflect the alignment between the current view direction and the Gaussian's preferred direction:
\begin{align}
f_{\text{cond}} &= \exp\left(-\lambda\,(d-\mu_d)^\top \Sigma_d^{-1}(d-\mu_d)\right), \\
\alpha_{\text{cond}} &= \alpha \cdot f_{\text{cond}},
\end{align}
where \(\lambda\) is a positive scaling parameter controlling the sensitivity of the modulation. This mechanism enhances the model’s ability to capture view-dependent effects such as specular highlights and anisotropic reflections. However, note that both 3DGS and 6DGS are inherently designed for static scenes, as they do not incorporate temporal dynamics.

\section{Our Approach}

\begin{figure*}[t]
\includegraphics[width=1.\textwidth]{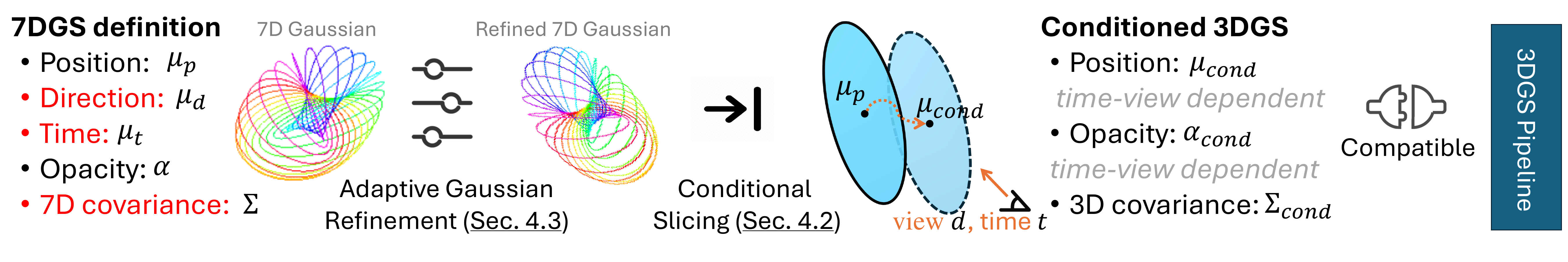}
\vspace{-2.5em}
\caption{Proposed 7DGS compatible with the existing 3DGS pipeline.}
\label{fig:pipeline}
\vspace{-1em}
\end{figure*}

We introduce 7D Gaussian Splatting (7DGS), a unified framework that jointly models spatial, temporal, and angular dimensions. In 7DGS, each scene element is represented as a 7D Gaussian that naturally captures scene geometry, dynamics, and view-dependent appearance. By extending the Gaussian representation with an additional temporal dimension, 7DGS seamlessly integrates spatial, temporal, and angular variations, preserving the advantages of efficient real-time rendering and accurate view-dependent effects while robustly handling dynamic scenes.

\subsection{7D Gaussian Representation}
In 7DGS, each scene element is modeled as a 7D Gaussian random variable that jointly encodes its spatial, temporal, and directional properties. This unified representation naturally captures not only the geometry of the scene but also its dynamics and view-dependent appearance. Formally, we define the 7D Gaussian as follows:
\begin{align}
\begin{small}
X = \begin{pmatrix} X_p \\ X_t \\ X_d \end{pmatrix} \sim \mathcal{N}\!\left(
\begin{pmatrix}
\mu_p \\ \mu_t \\ \mu_d
\end{pmatrix},
\begin{pmatrix}
\Sigma_p & \Sigma_{pt} & \Sigma_{pd} \\
\Sigma_{pt}^\top & \Sigma_t & \Sigma_{td} \\
\Sigma_{pd}^\top & \Sigma_{td}^\top & \Sigma_d
\end{pmatrix}
\right),
\end{small}
\end{align}
where:
\begin{itemize}
    \item \(X_p \in \mathbb{R}^3\) represents the spatial coordinates, with mean \(\mu_p\) and covariance \(\Sigma_p\) that model the local geometric shape.
    \item \(X_t \in \mathbb{R}\) is a scalar capturing the temporal coordinate, with mean \(\mu_t\) and variance \(\Sigma_t\). This component accounts for the dynamic evolution of scene elements.
    \item \(X_d \in \mathbb{R}^3\) encodes the directional (angular) information, with mean \(\mu_d\) and covariance \(\Sigma_d\), which is critical for modeling view-dependent effects.
\end{itemize}
The off-diagonal blocks \(\Sigma_{pt}\), \(\Sigma_{pd}\), and \(\Sigma_{td}\) capture the correlations among the spatial, temporal, and directional components, enabling the Gaussian to model complex interdependencies across these dimensions.

Inspired by 6DGS, we parameterize the full 7D covariance matrix using a Cholesky decomposition:
\begin{align}
  \Sigma = LL^\top,
\end{align}
where \(L\) is a lower-triangular matrix with positive diagonal entries. This reparameterization not only guarantees a valid covariance matrix during optimization but also facilitates efficient computation.

For the color representation, we continue to adopt the view-dependent spherical harmonics formulation from 3DGS without introducing additional temporal dependencies, as the dynamic information is already encoded within the Gaussian parameters.

\subsection{Conditional Slicing Mechanism}
\label{sec:slice}
To render an image at a specified time \(t\) and from a given view direction \(d\), we condition each 7D Gaussian on the observed temporal and angular values. This operation “slices” the full 7D Gaussian to yield a conditional 3D Gaussian that solely governs the spatial component. Such conditioning is critical because it allows us to efficiently integrate the temporal dynamics and view-dependent effects into the traditional 3D rendering pipeline.

We begin by partitioning the covariance matrix into two parts: \(\Sigma_{(t,d)}\), corresponds to the temporal and directional dimensions, while the other, \(\Sigma_{p,(t,d)}\), links the spatial dimension with the combined temporal-directional space:
\[
\Sigma_{(t,d)} = \begin{pmatrix} \Sigma_t & \Sigma_{td} \\ \Sigma_{td}^\top & \Sigma_d \end{pmatrix}, \quad \text{and} \quad
\Sigma_{p,(t,d)} = \begin{bmatrix} \Sigma_{pt} & \Sigma_{pd} \end{bmatrix}.
\]
Here, \(\Sigma_t\) and \(\Sigma_d\) are the covariance matrices associated with the temporal and directional components, respectively, and \(\Sigma_{td}\) captures their mutual correlation. Similarly, \(\Sigma_{pt}\) and \(\Sigma_{pd}\) encode how the spatial component correlates with time and view direction.

Using the standard properties of multivariate Gaussian distributions, the conditional distribution of the spatial component \(X_p\) given \(X_t = t\) and \(X_d = d\) is also Gaussian:
\begin{align}
p(X_p \mid X_t = t, \, X_d = d) \sim \mathcal{N}(\mu_{\text{cond}}, \Sigma_{\text{cond}}),
\end{align}
with conditional mean and covariance given by
\begin{align}
\mu_{\text{cond}} &= \mu_p + \Sigma_{p,(t,d)} \, \Sigma_{(t,d)}^{-1} \, \begin{pmatrix} t-\mu_t \\ d-\mu_d \end{pmatrix}, \label{eq:cond_mean} \\
\Sigma_{\text{cond}} &= \Sigma_p - \Sigma_{p,(t,d)} \, \Sigma_{(t,d)}^{-1} \, \Sigma_{p,(t,d)}^\top. \label{eq:cond_cov}
\end{align}
In Equation (\ref{eq:cond_mean}), the term \(\Sigma_{p,(t,d)} \, \Sigma_{(t,d)}^{-1} \, \begin{pmatrix} t-\mu_t \\ d-\mu_d \end{pmatrix}\) serves as a correction that shifts the spatial mean \(\mu_p\) in accordance with deviations in time and view direction from their expected values \(\mu_t\) and \(\mu_d\). Equation (\ref{eq:cond_cov}) similarly adjusts the spatial uncertainty \(\Sigma_p\) by removing the part of the variance explained by the temporal and directional components.

To further refine the rendering, we modulate the contribution of each Gaussian based on how much the observed time \(t\) and view direction \(d\) deviate from the Gaussian’s expected values. We define two separate modulation factors:
\begin{align}
f_{\text{temp}} &= \exp\!\left( -\frac{1}{2}\,\lambda_t\,(t-\mu_t)^2\,\Sigma_t^{-1} \right), \label{eq:temp_mod} \\
f_{\text{dir}} &= \exp\!\left( -\frac{1}{2}\,\lambda_d\,(d-\mu_d)^\top \Sigma_d^{-1}(d-\mu_d) \right), \label{eq:dir_mod}
\end{align}
where \(\lambda_t\) and \(\lambda_d\) are positive scalar parameters that control the sensitivity of the temporal and directional modulation, respectively. The factor \(f_{\text{temp}}\) decays exponentially as the observed time \(t\) diverges from the expected time \(\mu_t\), with the decay rate governed by \(\lambda_t\). Similarly, the factor \(f_{\text{dir}}\) decreases as the view direction \(d\) moves away from the preferred direction \(\mu_d\).

The final conditional opacity for the Gaussian is then computed by combining the base opacity \(\alpha\) with both modulation factors:
\begin{align}
\alpha_{\text{cond}} = \alpha \cdot f_{\text{temp}} \cdot f_{\text{dir}}.
\end{align}
This formulation ensures that Gaussians contribute less to the rendered image when the current time or view direction is far from their expected values, thereby effectively integrating temporal dynamics and view-dependent appearance into the rendering process.

\subsection{Adaptive Gaussian Refinement}
While the conditional slicing mechanism in 7DGS adjusts the spatial mean \(\mu_{\text{cond}}\) and modulates the opacity \(\alpha_{\text{cond}}\) based on the current time \(t\) and view direction \(d\), the intrinsic shape of each Gaussian—determined by its covariance—remains static over time. This limitation can hinder representing complex dynamic behaviors such as non-rigid deformations or motion-induced shape changes. To address this, we introduce an \emph{Adaptive Gaussian Refinement} that dynamically updates the Gaussian parameters via residual corrections computed by lightweight neural networks.

Specifically, we first construct a comprehensive feature vector \(f\) that encapsulates the geometric and temporal context of each Gaussian. This feature vector is formed by concatenating the spatial mean \(\mu_p\), the temporal coordinate \(\mu_t\), the directional mean \(\mu_d\), and a high-frequency temporal encoding \(\gamma(t)\):
\begin{align}
    f &= \mu_p \oplus \mu_t \oplus \mu_d \oplus \gamma(t),
\end{align}
where \(\oplus\) denotes vector concatenation. The temporal encoding \(\gamma(t)\) is defined as
\begin{small}
\begin{align}
    \gamma(t) = \Bigl( \sin(2^0 \pi t),\, \cos(2^0 \pi t),\, \dots,\, \sin(2^{K-1} \pi t),\, \cos(2^{K-1} \pi t) \Bigr),\nonumber
\end{align}
\end{small}with \(K=10\). This multi-frequency encoding, inspired by positional encodings in \cite{mildenhall2020nerf}, provides a rich representation of time that captures both low-frequency trends and high-frequency details.

Next, we employ a set of small two-layer multilayer perceptrons (MLPs) with architecture \(C_{\text{in}} \times 64 \times C_{\text{out}}\) to predict residual adjustments for the key Gaussian parameters. These residuals are added to the original parameters to yield refined estimates:
\begin{equation}
\begin{array}{lll}
\hat{\mu}_p &= \mu_p + \phi_p(f), & \hat{\mu}_t = \mu_t + \phi_t(f), \\
\hat{\mu}_d &= \mu_d + \phi_d(f), & \hat{l}   = l + \phi_l(f).
\end{array}
\end{equation}
Here, \(l\) represents the vectorized lower-triangular elements of the 7D covariance matrix, and \(\phi_p(f)\), \(\phi_t(f)\), \(\phi_d(f)\), and \(\phi_l(f)\) are the residuals predicted by the respective MLPs. These updates allow the spatial position, temporal coordinate, directional mean, and covariance (which controls rotation and shape) to be dynamically adjusted as functions of the observed time.

This refinement module is applied before the conditional slicing step (Section \ref{sec:slice}). By dynamically adapting the 7D Gaussian parameters, the subsequent conditioning produces a 3D Gaussian whose spatial attributes—including its shape and orientation—more accurately reflect the evolving scene dynamics and view-dependent variations. This leads to improved modeling of complex motions and a more faithful reconstruction of dynamic scenes.

\begingroup
\begin{algorithm}[t]
\centering
\caption{Slice 7DGS to Conditional 3DGS}
\label{al:slice}
\renewcommand{\algorithmicrequire}{\textbf{Input:}}
\renewcommand{\algorithmicensure}{\textbf{Output:}}
\begin{algorithmic}[1]
\Require Lower-triangular \(L\), \(\mu_p\), \(\mu_t\), \(\mu_d\), base opacity \(\alpha\), scaling factors \(\lambda_{t}, \lambda_{d}\), view direction \(d\), observed time \(t\)
\Ensure Conditional \(\mu_{\text{cond}}\), \(\Sigma_{\text{cond}}\), \(\alpha_{\text{cond}}\) (optionally, scale \(S\) and rotation \(R\) are required for densification steps)
\State Compute feature: \(f = \text{concat}\bigl(\mu_p,\, \mu_t,\, \mu_d,\, \gamma(t)\bigr)\).
\State Adaptive Gaussian refinement:
\[
\begin{array}{lll}
&\hat{\mu}_p = \mu_p + \phi_p(f), \quad &\hat{\mu}_t = \mu_t + \phi_t(f),  \\
&\hat{\mu}_d = \mu_d + \phi_d(f),  \quad  &\hat{l}   = l + \phi_l(f).
\end{array}
\]
\State Reconstruct refined covariance: \(\hat{\Sigma} = \hat{L}\hat{L}^\top\).
\State Partition \(\hat{\Sigma}\) into blocks:
\[
\hat{\Sigma} =
\begin{pmatrix}
\hat{\Sigma}_p & \hat{\Sigma}_{p,(t,d)} \\
\hat{\Sigma}_{p,(t,d)}^\top & \hat{\Sigma}_{(t,d)}
\end{pmatrix}, \text{with}\; \hat{\Sigma}_{(t,d)} = \begin{pmatrix}\hat{\Sigma}_t & \hat{\Sigma}_{td} \\ \hat{\Sigma}_{td}^\top & \hat{\Sigma}_d\end{pmatrix}\].
\State Compute conditional statistics:
\[
\begin{array}{rcl}
\Sigma_{\text{cond}} &=& \hat{\Sigma}_p - \hat{\Sigma}_{p,(t,d)}\,\hat{\Sigma}_{(t,d)}^{-1}\,\hat{\Sigma}_{p,(t,d)}^\top,\\
\mu_{\text{cond}} &=& \hat{\mu}_p + \hat{\Sigma}_{p,(t,d)}\,\hat{\Sigma}_{(t,d)}^{-1}
\begin{pmatrix}t-\hat{\mu}_t\\ d-\hat{\mu}_d\end{pmatrix}.
\end{array}
\]
\State Compute conditional opacity:
\[
\begin{array}{rcl}
f_{\text{temp}} &=& \exp\!\left(-\frac{1}{2}\,\lambda_t\,(t-\hat{\mu}_t)^2\,\hat{\Sigma}_t^{-1}\right),\\
f_{\text{dir}}  &=& \exp\!\left(-\frac{1}{2}\,\lambda_d\,(d-\hat{\mu}_d)^\top\hat{\Sigma}_d^{-1}(d-\hat{\mu}_d)\right),\\
\alpha_{\text{cond}} &=& \alpha \cdot f_{\text{temp}} \cdot f_{\text{dir}}
\end{array}
\]
\State \textbf{Optional:} Perform SVD on \(\Sigma_{\text{cond}} = UDU^\top\) to extract scale \(S=\sqrt{\operatorname{diag}(D)}\) and rotation \(R=U\) (adjust \(R\) to ensure \(\det(R)>0\)) on desification steps.
\end{algorithmic}
\end{algorithm}
\endgroup

\subsection{Optimization and Rendering Pipeline}

Our optimization strategy extends the adaptive Gaussian densification framework of 3DGS to the enriched spatio-temporal-angular domain of 7DGS. In our method, each Gaussian is dynamically adjusted via cloning and splitting operations, ensuring comprehensive coverage across spatial, temporal, and directional dimensions.

To guide these refinement operations, we first extract scale and rotation information from the conditional covariance matrix \(\Sigma_{\text{cond}}\) (obtained after conditioning on the observed time \(t\) and view direction \(d\)). We perform a Singular Value Decomposition:
\(
\Sigma_{\text{cond}} = U\,D\,U^\top,
\)
where \(U\) is an orthogonal matrix and \(D\) is a diagonal matrix containing the singular values. We then define the rotation matrix as \(R = U\) and compute the scale vector as
\(
S = \sqrt{\operatorname{diag}(D)}.
\)
To ensure that \(R\) represents a right-handed coordinate system, we adjust its last column as follows:
\(
R_{:,3} = R_{:,3} \cdot \operatorname{sign}(\det(R)).
\)

For Gaussian splitting, 7DGS leverages temporal cues in addition to spatial gradients. We quantify the spatial-temporal correlation using the magnitude of the off-diagonal block \(\Sigma_{pt}\), which captures the interaction between spatial and temporal components. When this correlation exceeds a threshold of 0.05 (relative to the screen extent) and the normalized temporal scale (derived from \(\Sigma_t\)) is larger than 0.25, the corresponding Gaussians are split. This criterion ensures that regions with significant motion dynamics are densely represented.

The rendering pipeline remains fully compatible with 3DGS. In our approach, the 7DGS representation is first converted into a 3DGS-compatible format via the conditional slicing mechanism (see Section \ref{sec:slice} and Algorithm \ref{al:slice}). The resulting conditional 3D Gaussian's mean and covariance are then projected onto the image plane using standard perspective projection, yielding a set of 2D Gaussians. These 2D Gaussians are subsequently splatted onto the image canvas using a differentiable rasterization routine, and the final pixel colors are computed by aggregating the contributions of all Gaussians in a depth-aware, opacity-blended manner.

Importantly, our 7DGS framework integrates seamlessly with the existing 3DGS training pipeline. We employ the same loss functions, optimizers, and hyperparameter settings—with the only modification being an increased minimum opacity threshold (\(\tau_{\text{min}}=0.01\)) for pruning, which compensates for the modulation of the conditional opacity \(\alpha_{\text{cond}}\) by time and view direction. By converting our 7D representation into a conditional 3D format, we fully leverage the adaptive density control and efficient rasterization techniques of 3DGS, thereby achieving enhanced performance with minimal modifications.

\section{Experiments}
\subsection{Experimental Protocol}
\begin{table*}[tbp]
  \centering
  \resizebox{\linewidth}{!}{
        \begin{tabular}{c|l|rrrrrr|rrrrrr||rrrrr}
    \toprule
    \multirow{2}[4]{*}{Dataset} & \multicolumn{1}{c|}{\multirow{2}[4]{*}{Scene}} & \multicolumn{6}{c|}{4DGS}                     & \multicolumn{6}{c||}{7DGS (\textbf{Ours})}              & \multicolumn{5}{c}{7DGS (\textbf{Ours}, w/o AGR)} \\
\cmidrule{3-19}          &       & \multicolumn{1}{c}{PSNR$\uparrow$} & \multicolumn{1}{c}{SSIM$\uparrow$} & \multicolumn{1}{c}{LPIPS$\downarrow$} & \multicolumn{1}{c}{Train$\downarrow$} & \multicolumn{1}{c}{FPS$\uparrow$} & \multicolumn{1}{c|}{\# points$\downarrow$} & \multicolumn{1}{c}{PSNR$\uparrow$} & \multicolumn{1}{c}{SSIM} & \multicolumn{1}{c}{LPIPS$\downarrow$} & \multicolumn{1}{c}{Train$\downarrow$} & \multicolumn{1}{c}{FPS$\uparrow$} & \multicolumn{1}{c||}{\# points$\downarrow$} & \multicolumn{1}{c}{PSNR$\uparrow$} & \multicolumn{1}{c}{SSIM$\uparrow$} & \multicolumn{1}{c}{LPIPS$\downarrow$} & \multicolumn{1}{c}{FPS$\uparrow$} & \multicolumn{1}{c}{\# points$\downarrow$} \\
    \midrule
    \multicolumn{1}{c|}{\multirow{7}[4]{*}{\begin{sideways}7DGS-PBR\end{sideways}}} & \texttt{heart1} & 27.30 & 0.949 & 0.046 & \textbf{103.0} & 186.9 & 694,006 & \textbf{35.48} & \textbf{0.986} & \textbf{0.020} & 114.2 & 155.7 & 82,813 & 34.66 & 0.983 & 0.023 & \textbf{401.0} & \textbf{82,412} \\
          & \texttt{heart2} & 25.13 & 0.920 & 0.084 & \textbf{103.4} & 160.4 & 869,245 & \textbf{31.80} & \textbf{0.964} & \textbf{0.051} & 145.9 & 139.5 & 98,458 & 30.99 & 0.959 & 0.057 & \textbf{384.6} & \textbf{101,503} \\
          & \texttt{cloud} & 24.63 & 0.938 & 0.100 & 123.7 & 219.0 & 216,878 & \textbf{29.60} & \textbf{0.955} & \textbf{0.075} & \textbf{102.6} & 199.2 & 44,858 & 29.29 & 0.955 & 0.075 & \textbf{386.2} & \textbf{44,175} \\
          & \texttt{dust} & 35.88 & 0.954 & 0.037 & 97.0  & 296.1 & 357,744 & \textbf{37.30} & \textbf{0.956} & \textbf{0.037} & \textbf{69.8} & 243.9 & 11,253 & 36.87 & 0.955 & 0.038 & \textbf{394.8} & \textbf{10,924} \\
          & \texttt{flame} & 29.34 & 0.928 & 0.067 & 113.7 & 151.2 & 947,786 & \textbf{32.53} & \textbf{0.940} & \textbf{0.059} & \textbf{74.1} & 247.2 & 16,544 & 31.67 & 0.937 & 0.062 & \textbf{371.6} & \textbf{15,060} \\
          & \texttt{suzanne} & 24.45 & 0.917 & 0.141 & 222.5 & 141.8 & 766,098 & \textbf{28.26} & \textbf{0.949} & \textbf{0.062} & \textbf{193.9} & 62.5  & 336,713 & 27.14 & 0.942 & 0.074 & \textbf{317.9} & \textbf{276,281} \\
\cmidrule{2-19}          & \cellcolor[rgb]{ .949,  .949,  .949}\texttt{avg} & \cellcolor[rgb]{ .949,  .949,  .949}27.79 & \cellcolor[rgb]{ .949,  .949,  .949}0.934 & \cellcolor[rgb]{ .949,  .949,  .949}0.079 & \cellcolor[rgb]{ .949,  .949,  .949}127.2 & \cellcolor[rgb]{ .949,  .949,  .949}192.6 & \cellcolor[rgb]{ .949,  .949,  .949}641,960 & \cellcolor[rgb]{ .949,  .949,  .949}\textbf{32.50} & \cellcolor[rgb]{ .949,  .949,  .949}\textbf{0.958} & \cellcolor[rgb]{ .949,  .949,  .949}\textbf{0.051} & \cellcolor[rgb]{ .949,  .949,  .949}\textbf{116.7} & \cellcolor[rgb]{ .949,  .949,  .949}174.7 & \cellcolor[rgb]{ .949,  .949,  .949}98,440 & \cellcolor[rgb]{ .949,  .949,  .949}31.77 & \cellcolor[rgb]{ .949,  .949,  .949}0.955 & \cellcolor[rgb]{ .949,  .949,  .949}0.055 & \cellcolor[rgb]{ .949,  .949,  .949}\textbf{376.0} & \cellcolor[rgb]{ .949,  .949,  .949}\textbf{88,393} \\
    \midrule
    \multicolumn{1}{c|}{\multirow{9}[4]{*}{\begin{sideways}D-NeRF\end{sideways}}} & \texttt{b.balls} & 33.24 & 0.982 & 0.025 & \textbf{50.7} & 219.9 & 276,073 & \textbf{35.10} & \textbf{0.984} & \textbf{0.019} & 86.6  & 104.3 & 129,791 & 34.05 & 0.982 & 0.025 & \textbf{213.1} & \textbf{127,395} \\
          & \texttt{h.warrior} & \textbf{34.10} & \textbf{0.949} & \textbf{0.067} & 35.3  & 299.4 & 298,391 & 32.96 & 0.935 & 0.084 & \textbf{31.3} & 238.3 & 9,569 & 32.78 & 0.934 & 0.090 & \textbf{431.5} & \textbf{8,693} \\
          & \texttt{hook} & \textbf{32.93} & \textbf{0.970} & \textbf{0.034} & 38.0  & 325.1 & 174,720 & 31.57 & 0.962 & 0.040 & \textbf{35.7} & 233.1 & 24,700 & 30.95 & 0.958 & 0.045 & \textbf{432.8} & \textbf{21,662} \\
          & \texttt{j.jacks} & 31.14 & 0.970 & 0.044 & 66.9  & 366.0 & 143,665 & \textbf{33.57} & \textbf{0.977} & \textbf{0.027} & \textbf{34.1} & 243.2 & 18,784 & 31.37 & 0.967 & 0.042 & \textbf{432.4} & \textbf{15,779} \\
          & \texttt{lego} & 25.58 & 0.917 & 0.077 & \textbf{55.4} & 320.0 & 186,165 & \textbf{28.86} & \textbf{0.947} & \textbf{0.051} & 78.5  & 160.2 & 74,884 & 28.72 & 0.947 & 0.051 & \textbf{365.0} & \textbf{68,552} \\
          & \texttt{mutant} & 39.01 & 0.991 & 0.009 & \textbf{39.1} & 341.6 & 138,691 & \textbf{41.36} & \textbf{0.995} & \textbf{0.005} & 42.5  & 193.7 & 37,706 & 39.59 & 0.993 & 0.007 & \textbf{395.8} & \textbf{33,868} \\
          & \texttt{standup} & 39.75 & 0.991 & 0.008 & 34.4  & 330.4 & 142,468 & \textbf{40.60} & \textbf{0.992} & \textbf{0.008} & \textbf{33.5} & 224.0 & 15,598 & 38.45 & 0.988 & 0.014 & \textbf{399.2} & \textbf{12,688} \\
          & \texttt{trex} & 29.89 & 0.979 & 0.021 & 100.7 & 169.2 & 682,378 & \textbf{30.72} & \textbf{0.980} & \textbf{0.018} & \textbf{63.4} & 156.5 & 67,994 & 30.13 & 0.979 & 0.021 & \textbf{352.4} & \textbf{61,946} \\
\cmidrule{2-19}          & \cellcolor[rgb]{ .949,  .949,  .949}\texttt{avg} & \cellcolor[rgb]{ .949,  .949,  .949}33.21 & \cellcolor[rgb]{ .949,  .949,  .949}0.969 & \cellcolor[rgb]{ .949,  .949,  .949}0.036 & \cellcolor[rgb]{ .949,  .949,  .949}52.6 & \cellcolor[rgb]{ .949,  .949,  .949}296.4 & \cellcolor[rgb]{ .949,  .949,  .949}255,319 & \cellcolor[rgb]{ .949,  .949,  .949}\textbf{34.34} & \cellcolor[rgb]{ .949,  .949,  .949}\textbf{0.972} & \cellcolor[rgb]{ .949,  .949,  .949}\textbf{0.032} & \cellcolor[rgb]{ .949,  .949,  .949}\textbf{50.7} & \cellcolor[rgb]{ .949,  .949,  .949}194.2 & \cellcolor[rgb]{ .949,  .949,  .949}47,378 & \cellcolor[rgb]{ .949,  .949,  .949}33.26 & \cellcolor[rgb]{ .949,  .949,  .949}0.969 & \cellcolor[rgb]{ .949,  .949,  .949}0.037 & \cellcolor[rgb]{ .949,  .949,  .949}\textbf{377.8} & \cellcolor[rgb]{ .949,  .949,  .949}\textbf{43,823} \\
    \midrule
    \multicolumn{1}{c|}{\multirow{6}[3]{*}{\begin{sideways}Technicolor\end{sideways}}} & \texttt{birthday} & 31.28 & 0.922 & 0.153 & 370.8 & 69.6  & 842,491 & \textbf{32.31} & \textbf{0.940} & \textbf{0.111} & \textbf{117.0} & 39.2  & 589,128 & 32.01 & 0.937 & 0.116 & \textbf{237.1} & \textbf{622,437} \\
          & \texttt{fabien} & \textbf{35.48} & \textbf{0.894} & \textbf{0.297} & 332.8 & 110.5 & 705,106 & 34.87 & 0.885 & 0.317 & \textbf{73.0} & 131.3 & 107,240 & 34.53 & 0.876 & 0.336 & \textbf{354.7} & \textbf{91,237} \\
          & \texttt{painter} & 35.09 & 0.905 & 0.238 & 361.9 & 99.4  & 287,036 & \textbf{36.54} & \textbf{0.919} & \textbf{0.208} & \textbf{98.1} & 113.1 & 144,226 & 36.46 & 0.914 & 0.216 & \textbf{364.0} & \textbf{119,654} \\
          & \texttt{theater} & \textbf{31.84} & 0.871 & 0.291 & 383.7 & 83.0  & 946,909 & 31.54 & \textbf{0.876} & \textbf{0.265} & \textbf{87.3} & 94.0  & 197,713 & 31.09 & 0.873 & 0.271 & \textbf{333.2} & \textbf{185,330} \\
          & \texttt{train} & 32.58 & 0.932 & 0.102 & 345.3 & 58.3  & 1,412,917 & \textbf{32.64} & \textbf{0.940} & \textbf{0.089} & \textbf{185.2} & 18.6  & 1,043,645 & 32.43 & 0.938 & 0.090 & \textbf{100.7} & \textbf{1,014,529} \\
\cmidrule{2-19}          & \cellcolor[rgb]{ .949,  .949,  .949}\texttt{avg} & \cellcolor[rgb]{ .949,  .949,  .949}33.25 & \cellcolor[rgb]{ .949,  .949,  .949}0.905 & \cellcolor[rgb]{ .949,  .949,  .949}0.216 & \cellcolor[rgb]{ .949,  .949,  .949}358.9 & \cellcolor[rgb]{ .949,  .949,  .949}84.2 & \cellcolor[rgb]{ .949,  .949,  .949}838,892 & \cellcolor[rgb]{ .949,  .949,  .949}\textbf{33.58} & \cellcolor[rgb]{ .949,  .949,  .949}\textbf{0.912} & \cellcolor[rgb]{ .949,  .949,  .949}\textbf{0.198} & \cellcolor[rgb]{ .949,  .949,  .949}\textbf{112.1} & \cellcolor[rgb]{ .949,  .949,  .949}79.2 & \cellcolor[rgb]{ .949,  .949,  .949}416,390 & \cellcolor[rgb]{ .949,  .949,  .949}33.30 & \cellcolor[rgb]{ .949,  .949,  .949}0.908 & \cellcolor[rgb]{ .949,  .949,  .949}0.206 & \cellcolor[rgb]{ .949,  .949,  .949}\textbf{278.0} & \cellcolor[rgb]{ .949,  .949,  .949}\textbf{406,637} \\
\bottomrule
    \end{tabular}}%
    \caption{Comparison with 4DGS \cite{yang2023real} on 7DGS-PBR, D-NeRF \cite{pumarola2021d}, and Technicolor \cite{sabater2017dataset}.`\textit{Train}' means training time in minutes.
    }
  \label{tab:result}%
\end{table*}%

\begin{figure*}[t]
\centering
\includegraphics[width=\linewidth]{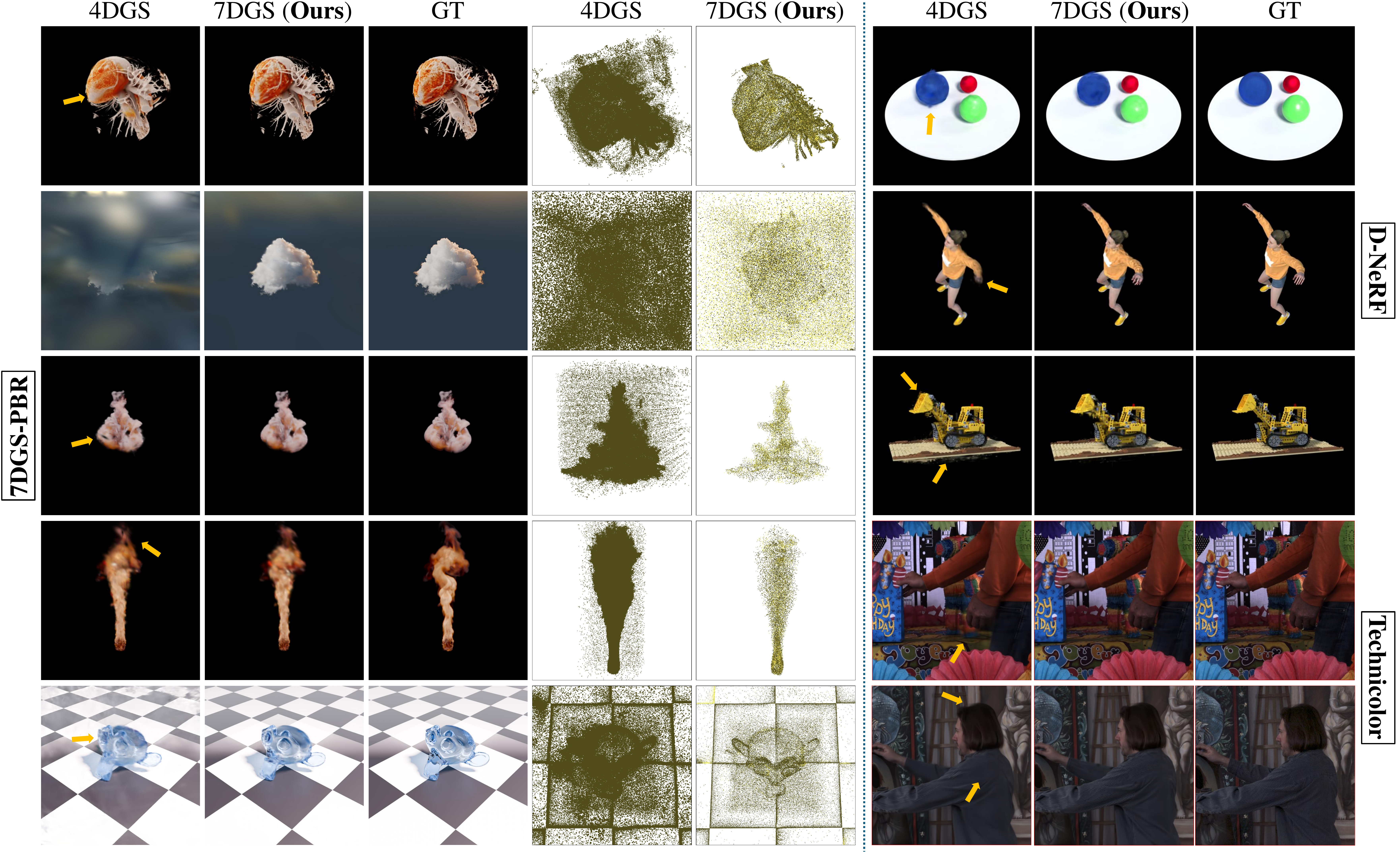}
\caption{Qualitative comparison of methods on the 7DGS-PBR, D-NeRF \cite{pumarola2021d}, and Technicolor \cite{sabater2017dataset} datasets (zoom in for details).}
\vspace{-1em}
\label{fig:qualitative}
\end{figure*}

\noindent\textbf{Datasets.}
We evaluate 7DGS on three distinct datasets:
\begin{itemize}
    \item \textbf{D-NeRF} \cite{pumarola2021d}: A synthetic monocular video dataset containing eight scenes at a resolution of $800\times800$.
    \item \textbf{Technicolor} \cite{sabater2017dataset}: An in-the-wild dataset composed of video recordings captured by a synchronized 4×4 camera array at $2048\times1088$ resolution.
    \item \textbf{7DGS-PBR}: Our custom dataset, rendered using physically-based techniques, consists of six dynamic scenes exhibiting complex view-dependent effects:
    \begin{itemize}
        \item \texttt{heart1} and \texttt{heart2}: Derived from real CT scans, these scenes capture cardiac cycles over 15 timestamps.
        \item \texttt{cloud}: Based on the Walt Disney Animation Studios volumetric cloud dataset\footnote{\href{https://disneyanimation.com/resources/clouds/}{Disney Animation Studios cloud dataset}}, this scene features a complete daylight cycle spanning 60 timestamps.
        \item \texttt{dust} and \texttt{flame}: Sourced from Blender Market\footnote{\href{https://blendermarket.com/products/animated-vdb-pack}{Animated VDB pack dataset}}, these scenes present dynamic volumetric effects over 79 and 101 timestamps, respectively.
        \item \texttt{suzanne}: the standard Blender test mesh rendered with a translucent “\textit{Glass BSDF}” material, showing jelly-like deformations across 60 timestamps.
    \end{itemize}
\end{itemize}
For each timestamp, we sampled 300, 60, 20, 10, 10, and 10 views for \texttt{heart1}, \texttt{heart2}, \texttt{cloud}, \texttt{dust}, \texttt{flame}, and \texttt{suzanne}, respectively, following a 9:1 train-test split.

All scenes were rendered using Blender's Cycles engine at a resolution of 1000×1000 for \texttt{heart1} and 1600×1600 for the remaining scenes. Average rendering times per view on an NVIDIA Tesla V100 GPU were 8 seconds for \texttt{heart1}, 18 seconds for \texttt{heart2}, 311 seconds for \texttt{cloud}, 16 seconds for \texttt{dust}, 28 seconds for \texttt{flame}, and 26 seconds for \texttt{suzanne}. We will make the 7DGS-PBR dataset publicly available.

\noindent\textbf{Evaluation Metrics.} 
We evaluate our method using three image quality metrics: Peak Signal-to-Noise Ratio (PSNR), Structural Similarity Index (SSIM) \cite{wang2004image}, and LPIPS \cite{zhang2018perceptual}. For efficiency, we report the number of Gaussian points, rendering speed (FPS), and training time (minutes).


\noindent\textbf{Implementation Details.}
We adopt the 4DGS configuration with a batch size of 4 and downscaling factor of 2, except for Technicolor \cite{sabater2017dataset} where we use no downscaling and reduce batch size to 1 for 7DGS experiments. The directional and temporal modulation parameters ($\lambda_d$ and $\lambda_t$) are initially set to 0.5 for 7DGS-PBR, 0.05 for D-NeRF, and 0.1 for Technicolor, based on their respective directional and temporal dependencies, and become trainable after 15,000 iterations. Point clouds for \texttt{heart1} and \texttt{heart2} are initialized using marching cubes following DDGS \cite{gao2024ddgs}, while other scenes use 100,000 randomly initialized points within a bounding cube. For Technicolor, we initialize from COLMAP sparse reconstructions.

All experiments are conducted on a single NVIDIA Tesla V100 GPU (16GB memory) using the Adam optimizer \cite{kingma2014adam}. We employ distinct learning rates for different parameter groups: $2.5 \times 10^{-2}$ for temporal and directional means ($\mu_t$, $\mu_d$), $5 \times 10^{-2}$ for covariance diagonals, $1 \times 10^{-2}$ for lower triangular covariance elements, and $2 \times 10^{-4}$ for adaptive Gaussian network parameters (which become trainable after 3,000 iterations). All remaining parameters follow the default 3DGS learning rates.

\subsection{Comparison with Baseline}
Table~\ref{tab:result} presents a comprehensive comparison between our 7DGS framework and the state-of-the-art 4DGS method across all three datasets. We evaluate both the full 7DGS implementation and a variant without the adaptive Gaussian refinement (AGR) component to isolate the contribution of our core 7D representation.

Our 7DGS consistently outperforms 4DGS across all evaluation metrics and datasets. On 7DGS-PBR, which specifically targets complex view-dependent effects, our method achieves remarkable improvements with an average PSNR gain of +4.71 dB (from 27.79 dB to 32.50 dB) while utilizing only 15.3\% of the Gaussian points required by 4DGS (98,440 vs. 641,960). The most substantial improvement is observed on the \texttt{heart1} scene, where 7DGS delivers an impressive +8.18 dB PSNR increase while requiring only 11.9\% of the Gaussian points used by 4DGS. In addition, our method reduces the training time by an average of 8.3\% and can even further speed up by implementing our 7DGS slicing in Algorithm \ref{al:slice} in CUDA.

On the D-NeRF dataset \cite{pumarola2021d}, 7DGS maintains its superior performance with an average PSNR improvement of +1.13 dB while using only 18.6\% of the Gaussian points. Similarly, on the challenging in-the-wild Technicolor dataset \cite{sabater2017dataset}, 7DGS delivers superior results with an average PSNR gain of +0.33 dB while requiring approximately half the number of Gaussian points.

Notably, even without the adaptive Gaussian refinement (AGR), our 7DGS (w/o AGR) variant still outperforms 4DGS with an average PSNR gain of +3.98 dB on 7DGS-PBR, +0.05 dB on D-NeRF, and +0.05 dB on Technicolor, while using significantly fewer Gaussian points (13.8\%, 17.2\%, and 48.5\% respectively). Additionally, the removal of AGR substantially accelerates rendering speed, achieving an average of 376.0 FPS, 377.8 FPS, and 278.0 FPS on the three datasets—approximately twice the rendering speed of the full 7DGS implementation and substantially faster than 4DGS.

Figure~\ref{fig:qualitative} provides visual comparisons of novel view renderings alongside visualizations of the reconstructed point clouds. The qualitative results reveal that 4DGS exhibits more pronounced artifacts, particularly for scenes with complex view-dependent effects. Furthermores, 7DGS produces cleaner, more faithful geometric reconstructions with superior handling of temporal dynamics and view-dependent appearance variations. The improvement is especially noticeable in scenes with complex lighting interactions, such as the translucent \texttt{suzanne} and the volumetric \texttt{cloud} scenes, where our unified spatio-temporal-angular representation effectively captures the interdependence between geometry, motion, and appearance.

\subsection{Comparison with State-of-the-Art}
\begin{table}[tbp]
  \centering
  \resizebox{1.0\linewidth}{!}{
    \begin{tabular}{c|l|rrr}
    \toprule
    \multirow{8}[4]{*}{\begin{sideways}D-NeRF\end{sideways}} & Method & \multicolumn{1}{c}{PSNR$\uparrow$} & \multicolumn{1}{c}{SSIM$\uparrow$} & \multicolumn{1}{c}{LPIPS$\downarrow$} \\
\cmidrule{2-5}          & D-NeRF \cite{pumarola2021d} & 29.67 & 0.95  & 0.07 \\
          & HexPlane \cite{cao2023hexplane} & 31.05 & 0.97  & 0.04 \\
          & K-Planes \cite{fridovich2023k} & 31.61 & 0.97  & - \\
          & DaReNeRF \cite{lou2024darenerf} & 31.95 & 0.97  & \textbf{0.03} \\
          & 4DGS* \cite{yang2023real} & 33.21 & 0.97  & 0.04 \\
          & 4DGaussians \cite{wu20244d} & 33.30 & \textbf{0.98} & \textbf{0.03} \\
          & 7DGS (\textbf{Ours}) & \textbf{34.34} & 0.97  & \textbf{0.03} \\
    \midrule
    \multirow{8}[4]{*}{\begin{sideways}Technicolor\end{sideways}} & Method & \multicolumn{1}{c}{PSNR$\uparrow$} & \multicolumn{1}{c}{SSIM$\uparrow$} & \multicolumn{1}{c}{LPIPS-Alex$\downarrow$} \\
\cmidrule{2-5}          & DyNeRF \cite{li2022neural} & 31.80 & -     & 0.142 \\
          & HyperReel \cite{park2021hypernerf} & 32.73 & 0.906 & 0.109 \\
          & 4DGaussians \cite{wu20244d} & 30.79 & 0.843 & 0.178 \\
          & STG \cite{li2024spacetime}   & 33.23 & 0.912 &\textbf{0.085} \\
          & 4DGS* \cite{yang2023real} & 33.25 & 0.905 & 0.110 \\
          & Ex4DGS* \cite{lee2024fully} & 33.49 & \textbf{0.917} & 0.094 \\
          & 7DGS (\textbf{Ours}) & \textbf{33.58} & 0.912 & 0.101 \\
    \bottomrule
    \end{tabular}}%
    \caption{Comparison with SOTA methods on benchmarks. Methods with `*' are reproduced results with the official codes. Note, SOTA methods only have 2-digit precision for LPIPS in D-NeRF.
    }
  \label{tab:sota}%
  \vspace{-1em}
\end{table}%

Table~\ref{tab:sota} presents a comprehensive comparison between our method and other state-of-the-art approaches on the D-NeRF \cite{pumarola2021d} and Technicolor \cite{sabater2017dataset} datasets.  On the D-NeRF dataset, 7DGS substantially outperforms all existing methods in terms of PSNR, achieving a score of 34.34 dB, which represents a +1.04 dB improvement over 4DGaussians \cite{wu20244d} and +1.13 dB over 4DGS \cite{yang2023real}. While 7DGS achieves a competitive SSIM of 0.97 (matching 4DGS and DaReNeRF), 4DGaussians slightly leads with 0.98. For LPIPS, our method ties for best performance with DaReNeRF and 4DGaussians at 0.03.

For the challenging Technicolor dataset, which features in-the-wild multi-view videos, our method achieves state-of-the-art results with a PSNR of 33.58 dB. In terms of SSIM, our score of 0.912 is competitive with the best-performing Ex4DGS (0.917) and matches STG exactly. While our LPIPS score of 0.101 is slightly higher (worse) than STG's leading 0.085, it represents an improvement over several other methods including 4DGS (0.110) and 4DGaussians (0.178).

The performance advantages across both datasets demonstrate the effectiveness of our unified 7D representation in handling diverse dynamic scenes. Furthermore, our 7DGS is inherently flexible and can be integrated with complementary techniques to further enhance performance. For instance, while Ex4DGS \cite{lee2024fully} employs keyframe interpolation to explicitly model large-scale motion, similar strategies could be incorporated into 7DGS as future work. The modular nature of our approach allows for such extensions without compromising its core unified representation of spatial, temporal, and angular dimensions. This versatility positions 7DGS as not just a standalone improvement but as a fundamental advancement that can serve as a foundation for future research in dynamic scene rendering.

\section{Conclusion}
We present 7DGS, a novel framework that unifies spatial, temporal, and angular dimensions into a single 7D Gaussian representation for dynamic scene rendering. Our conditional slicing mechanism efficiently projects 7D Gaussians into renderable 3D Gaussians, enabling both high-quality results and real-time performance. Experiments across three datasets demonstrate that 7DGS outperforms state-of-the-art methods by up to 7.36 dB PSNR while using significantly fewer Gaussian points and maintaining render speeds of over 400 FPS (without adaptive refinement). Our approach excels particularly on scenes with complex view-dependent effects, advancing the field toward unified and efficient dynamic scene representations.

{
    \small
    \bibliographystyle{ieeenat_fullname}
    \bibliography{main}
}

\end{document}